\documentclass[a4paper]{llncs}

\usepackage{amssymb}
\usepackage{amsmath}
\usepackage{wrapfig}
\usepackage{graphicx}
\usepackage{multirow}
\usepackage[normalem]{ulem}
\useunder{\uline}{\ul}{}
\usepackage{bbding}
\usepackage{hyperref}
\usepackage{ifxetex}
\usepackage{ifluatex}
\usepackage{subfig}
\usepackage{tikz}
\usepackage{pgf}
\usepackage{pgfplots}
\usepackage[labelfont=bf]{caption}
\usepackage{color}
\usepackage[draft]{todonotes}   

\usetikzlibrary{positioning, shadows,arrows,automata}
\tikzset{
    state/.style={
           rectangle,
           rounded corners,
           draw=black, thick,
           minimum height=2em,
           inner sep=2pt,
           text centered,
           },
}

\definecolor{ashgrey}{rgb}{0.7, 0.75, 0.71}
\definecolor{aurometalsaurus}{rgb}{0.43, 0.5, 0.5}
\definecolor{arsenic}{rgb}{0.23, 0.27, 0.29}
\definecolor{battleshipgrey}{rgb}{0.52, 0.52, 0.51}
\definecolor{cadetgrey}{rgb}{0.57, 0.64, 0.69}
\definecolor{darkgray}{rgb}{0.66, 0.66, 0.66}
\definecolor{coolgrey}{rgb}{0.55, 0.57, 0.67}

\newcommand{\tit}{What Does Explainable AI Really Mean?\\
A New Conceptualization of Perspectives}

\begin{document}
\mainmatter  

\title{\tit}

\titlerunning{What Does Explainable AI Really Mean?}

\author{Derek Doran\inst{1} \and Sarah Schulz\inst{2} \and Tarek R. Besold\inst{3}}
\authorrunning{Doran et al.}

\institute{{Dept. of Computer Science \& Engineering, Kno.e.sis Research Center\\
Wright State University,
Dayton, Ohio, USA\\
\email{derek.doran@wright.edu}}
\and
Institute for Natural Language Processing\\
	University of Stuttgart, Stuttgart, Germany\\
	\email{sarah.schulz@ims.uni-stuttgart.de}
\and
Department of Computer Science\\
City, University of London, London, UK\\
\email{tarek-r.besold@city.ac.uk}}
\maketitle

\begin{abstract}
We characterize three notions of explainable AI that
cut across research fields: {\em opaque systems} that offer no insight into
its algorithmic mechanisms; {\em interpretable systems} where users can 
mathematically analyze its algorithmic mechanisms; and {\em comprehensible
systems} that emit symbols enabling user-driven explanations
of how a conclusion is reached. The paper is motivated by a corpus analysis of
NIPS, ACL, COGSCI, and ICCV/ECCV paper titles showing differences in how work on explainable
AI is positioned in various fields. We close by introducing a fourth notion: 
truly {\em explainable systems}, where automated reasoning is central to 
output crafted explanations without requiring human post processing as final step of the generative process. 
\end{abstract}

\section{Introduction}

If you were
held accountable for the decision of a machine in contexts that have financial, safety, security, or 
personal ramifications to an individual, would you blindly trust its decision?
How can we hold accountable Artificial Intelligence (AI) systems that make decisions
on possibly unethical grounds, e.g. when they predict a person's weight and health by their 
social media images~\cite{kocabey17} or the world region they are from~\cite{katti17}
as part of a downstream determination about their future, like when they
will quit their job~\cite{saradhi11}, commit a crime~\cite{gerber14}, or could be 
radicalized into terrorism~\cite{al06}? It is hard
to imagine a person who would feel comfortable in blindly agreeing with a system's decision 
in such highly 
consequential and ethical situations without a 
deep understanding of the decision making rationale of the system.
To achieve complete trustworthiness 
and an evaluation of the ethical and moral standards of a machine~\cite{skirpan17}, 
detailed ``explanations" of AI decisions seem necessary. 
Such explanations should provide insight into the rationale 
the AI uses to draw a conclusion. Yet many analysts indeed blindly `accept' 
the outcome of an AI, whether by necessity or by choice. To overcome this dangerous
practice, it is prudent for an AI to provide not only an output, but also a human 
understandable explanation that expresses the rationale of the machine. Analysts
can turn to such explanations to evaluate if a decision is reached by rational arguments
and does not incorporate reasoning steps conflicting with ethical or legal norms. 

But what constitutes an explanation? 
The Oxford English dictionary has no entry for the term `explainable', but has one for {\em explanation}:
{\em A statement or account that makes something clear; a {\bf reason} or justification given for an 
action or belief}. Do present systems that claim to make `explainable' decisions really provide 
explanations? Those who argue yes may point to Machine Learning (ML) algorithms that 
produce rules about data features to establish a classification decision, such as those learned
by decision trees~\cite{shafiq14}.
Others suggest that rich visualizations or text supplied along with a 
decision, as is often done in deep learning for computer vision~\cite{you16,johnson16,karpathy15},
offer sufficient information to draw an explanation of why a 
particular decision was reached. Yet ``rules" merely shed light
into {\em how}, not {\em why}, decisions are made, and supplementary artifacts of learning
systems (e.g. annotations and visualizations) require human-driven post processing under
their own line of reasoning. The variety of ways ``explanations" are currently handled is well 
summarized by Lipton~\cite{lipton16} when he states that {\em ``the term interpretability holds no agreed upon meaning, and yet machine learning conferences
frequently publish papers which wield the term in a quasi-mathematical
way''.} He goes on to call for engagement in the formulation of problems and 
their definitions to organize and advance explainable AI research.
In this position paper, we respond to Lipton's call by proposing various ``types" of explainable AI 
that cut across many fields of AI. 
\vspace{-12px}

\begin{wrapfigure}{l}{0.5\textwidth}
\centering
\vspace{-20px}
\scalebox{0.5}{
\begin{tikzpicture}
    \begin{axis}[
        scale only axis,
        ymin = 0,ymax=1.5,
        axis y line*=left,
        xlabel=\textsc{Year},
        xticklabel style=
{/pgf/number format/1000 sep=,rotate=60,anchor=east,font=\scriptsize},
        ylabel=\textsc{Term frequency},xtick=data,legend style={at={(1,0.8)},anchor=south east}
    ]
    
      \addplot[arsenic,mark=triangle] plot coordinates {
        (2007,     0.178757197908)
        (2008,     0.12644699011)
        (2009,     0.125832568329)
        (2010,     0.172279708273)
        (2011,     0.17524526912)
        (2012,    0.156473096407)
        (2013,    0.156195462478)
        (2014,    0.137747462325)
        (2015,    0.162977618617)
        (2016,    0.116903456032)
        
    };
    \addlegendentry{ACL}
        \addplot[cadetgrey,mark=square] plot coordinates {
        (2007,     1.21815784627)
        (2008,     1.40811550514)
        (2009,     1.03571770043)
        (2010,     1.07626620595)
        (2011,     1.05187946373)
        (2012,    1.18302183273)
        (2013,    1.18219945394)
        (2014,    1.04755090141)
        (2015,    1.17209487126)
        (2016,    0.94062016295)
    };
       \addlegendentry{COGSCI}
        \addplot[ashgrey,mark=o] plot coordinates {
        (2007,     0.183734136704)
        (2008,     0.246583307089)
        (2009,     0.194127378252)
        (2010,     0.192975685064)
        (2011,     0.17559262511)
        (2012,    0.17043127503)
        (2013,    0.15423951023)
        (2014,    0.123031090098)
        (2015,    0.130546430057)
        (2016,    0.163301393904)
    };
    \addlegendentry{NIPS}
        \addplot[darkgray,mark=diamond] plot coordinates {
        (2008,    0.128618681957)
        (2010,    0.179326023388)
        (2012,    0.16997903853)
        (2013,    0.144467431134)
        (2014,    0.154349199629)
        (2015,    0.127751727181)
        (2016,    0.118976799524)
    };
    \addlegendentry{ICCV/ECCV}
    
    \end{axis}
\end{tikzpicture}
}
\caption{Normalized corpus frequency of ``explain" or ``explanation".}
\label{fig:one}
\vspace{-25px}
\end{wrapfigure}
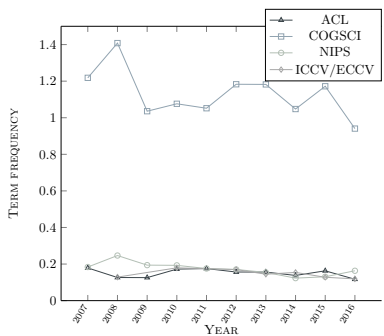

\section{Existing Perspectives in Explainable AI}
\vspace{-10px}
As stated by Lipton, terms like \textit{interpretability} are used in research papers, despite the lack of a clear and widely shared definition. In order to quantify this observation, we suggest a corpus-based analysis of relevant terms across research communities which strongly rely on ML-driven methodologies. The goal of this analysis is
to gain insights into the relevance of explainability across AI-related research communities and to detect how each fields defines notions of explainability. We carried out an experiment over
corpora of papers from the computer vision, NLP, connectionist, and symbolic reasoning communities.
We base our analysis on corpus statistics compiled from the proceedings of 
conferences where researchers employ, inter alia, ML techniques to approach their 
research objectives: the Annual Meeting of the Association for Computational Linguistics (ACL), the Annual Conference on Neural Information Processing Systems (NIPS), the International/European Conference on Computer Vision (ICCV/ECCV), and the Annual Conference of the Cognitive Science Society (COGSCI). The corpora include proceedings from 2007 to 2016. This allows us to observe trends regarding the use of words related to various concepts and the scope these concepts take.
We perform a shallow linguistic search for what we henceforth call ``explanation terms''. We apply simple substring matches such as ``explain'' or ``explanation'' for explainability, ``interpret'' for interpretability and ``compreh'' for comprehensibility. ``Explanation terms'' serve as an approximation to aspects of explainability. Frequencies are normalized, making them comparable between years and conferences.

\begin{figure}
	\includegraphics[width = \textwidth]{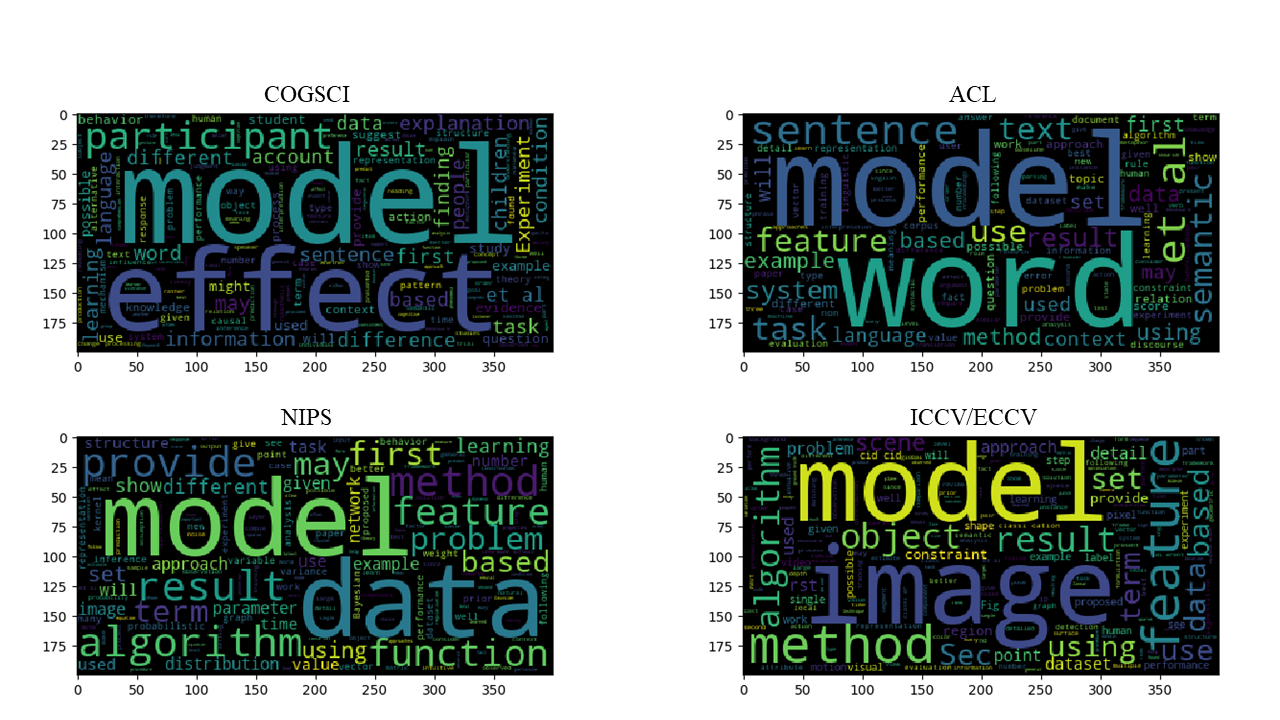}
	\caption{Word clouds of the context of explanation terms in the different proceedings corpora.}
	\label{fig:cloud}
	\vspace{-20px}
\end{figure}

The normalized frequencies for explanation terms are plotted in Figure~\ref{fig:one}. 
We omit frequency plots for interpretation and comprehension terms because they
exhibit a similar pattern. The frequency level of explainability concepts for COGSCI
is significantly above those of the other corpora. This could be 
due to the fact that Cognitive Science explicitly aims to explain the mind and its processes, in many cases leading to qualitatively different research questions, corresponding terminology, and actively used vocabularies. The NIPS corpus hints at an emphasis of explainability in 2008 and a slight increase in interest in this concept in 2016 in the connectionist community. 
To better understand how consistent topics and ideas around explainability are
across fields, we also analyze the context of its mentions. Word clouds shown in Figure~\ref{fig:cloud} are 
a simple method to gain an intuition about the composition of a concept and its semantic contents 
by highlighting the important words related to it. Important words are defined as words that appear within a 20 words window of a mention of an explanation term with a frequency highly above 
average\footnote{Word clouds are generated with the word-cloud package (\url{http://amueller.github.io/word_cloud/index.html}).}.

%

All communities focus on the explainability of a \textit{model} but there is a difference between the nature of models in Cognitive Science and the other fields. The COGSCI corpus mentions a \textit{participant}, a \textit{task} and an \textit{effect} whereas the other communities focus on a \textit{model} and what constitutes \textit{data} in their fields. It is not surprising that the neural modeling and NLP communities show a large overlap in their usage of explainability since there is an overlap in the research communities as well. We note further differences across the three ML communities compared to COGSCI. In the ACL corpus, explainability is often paired with words like features, examples, and words, which could suggest an emphasis on using examples to demonstrate the decision making of NLP decisions and the relevance of particular features. In 
the NIPS corpus, explainability is more closely tied to methods, algorithms, and results suggesting
a  desire to establish explanations about how neural systems translate inputs to outputs. The ICCV/ECCV falls 
between the ACL and NIPS corpus in the sense that it pairs explainability with
data (images) and features (objects) like ACL, but may also tie the notion to how
algorithms use (using) images to generate outputs. 

The corpus analysis establishes some differences in how
various AI communities approach the concept of explainability. In particular, 
we note that the term is sometimes used to help probe the mechanisms 
of ML systems (e.g. we seek an {\em interpretation} of how the system works), 
and other times to relate explanations to particular inputs and 
examples (e.g. we want to {\em comprehend} how an input was mapped to an output).
We use these observations to develop the following notions, also illustrated in Figure~\ref{fig:current}:

{\bf Opaque systems.} A system where the
	mechanisms mapping inputs to outputs are invisible 
	to the user. It can be seen as an oracle that makes predictions over an input, without indicating how and why predictions are made. Opaque systems emerge, for instance, when closed-source AI is licensed by an organization, where the licensor does not want to reveal the workings of its proprietary AI. Similarly, systems relying on genuine ``black box'' approaches, for which inspection of the algorithm or implementation does not give insight into the system's actual reasoning from inputs to corresponding outputs, are classified as opaque.
	
{\bf Interpretable systems.} A system where a user cannot only see, but also study and understand
    how inputs are mathematically mapped to outputs. 
	This implies model {\em transparency}, and requires 
    a level of understanding of the technical details of the mapping. 
	A regression model can be interpreted by comparing covariate weights to realize the relative importance of each feature to the mapping. SVMs and 
	other linear classifiers are interpretable insofar as data classes 
	are defined by their location relative to decision boundaries. But the action of deep neural networks, where input features may 
	be automatically learned and transformed through 
	non-linearities, is unlikely to be interpretable by most users.
	
{\bf Comprehensible systems.} A comprehensible system emits symbols along with its output (echoing Michie's {\em strong} and {\em ultra-strong machine learning} \cite{michie88}). These symbols (most often words, but also visualizations, etc.) allow
    the user to relate properties of the inputs to their output.
    The user is responsible for compiling and comprehending the symbols, relying on her own implicit form of knowledge and reasoning about them. This makes comprehensibility a graded notion, with the degree of a system's comprehensibility corresponding to the relative ease or difficulty of the compilation and comprehension. The required implicit form of knowledge on the side of the user is often 
	an implicit cognitive ``intuition'' about how the input, the symbols, and
	the output relate to each other. Taking the image in Figure~\ref{fig:current} as example, it is intuitive to think that users will comprehend the symbols by noting that they represent objects observed in the image, and that the objects may be related to each other as items often seen in a factory. Different users may have different tolerances in their comprehension: some may be willing to draw arbitrary relationships between objects while others would only be satisfied under a highly constrained set of assumptions.\\

\section{Defining Notions of Explainability}

\begin{figure}
\centering
\includegraphics[width=\textwidth]{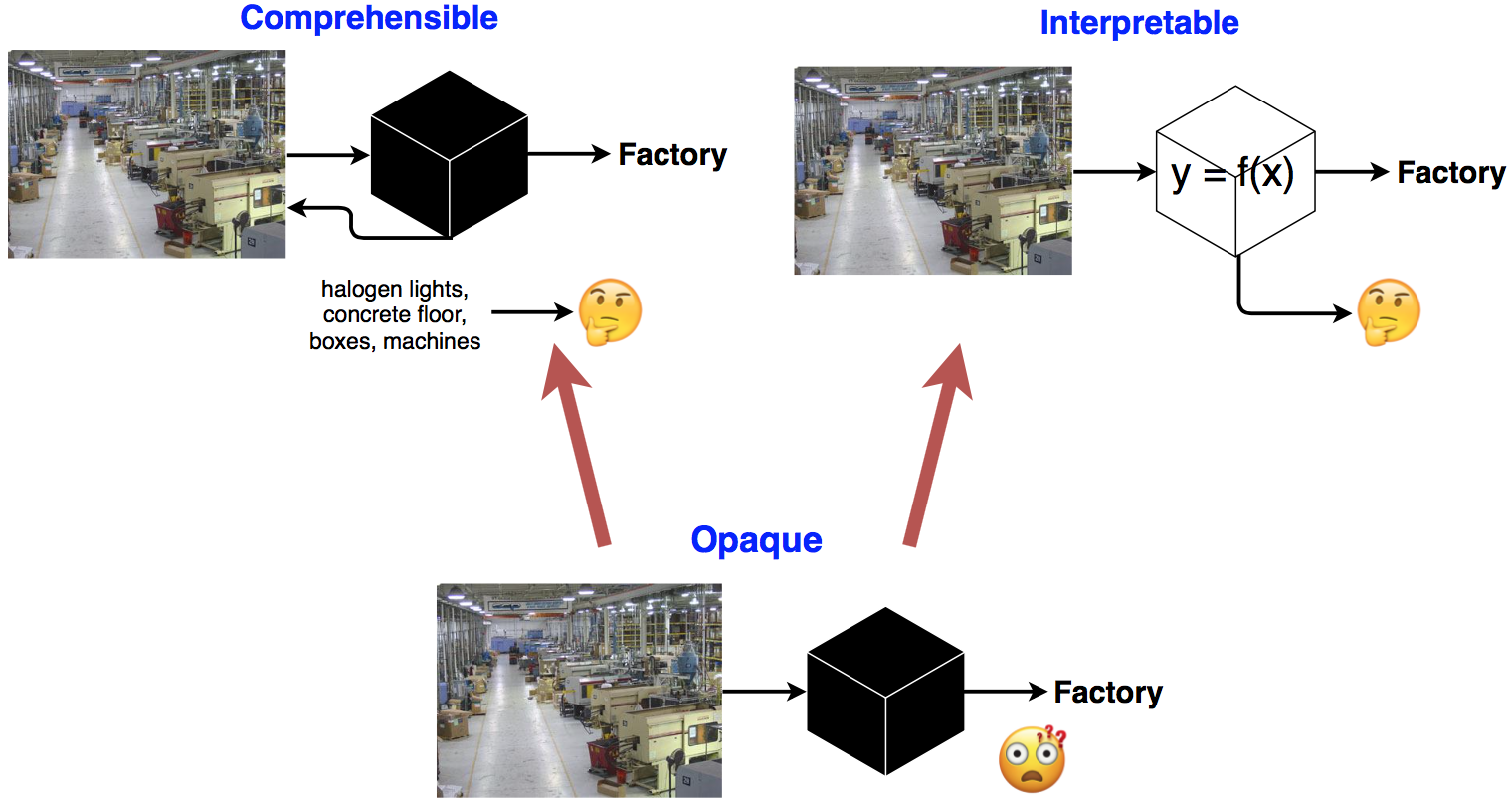}
\caption{Relation between opaque, comprehensible, and interpretable AI.} 
\label{fig:current}
\vspace{-15px}
\end{figure}

\noindent The arrows in Figure~\ref{fig:current} suggest that comprehensible and
interpretable systems each are improvements over opaque systems. The notions of comprehension and interpretation
are separate: while interpretation requires transparency in the underlying
mechanisms of a system, a comprehensible one can be opaque while emitting 
symbols a user can reason over. Regression models, support vector machines, 
decision trees, ANOVAs, and data clustering (assuming a kernel that is 
itself interpretable) are common examples of interpretable models. 
High dimensional data visualizations like t-SNE~\cite{maaten08} and 
receptive field visualization
on convolutional neural networks~\cite{bau17} are examples of 
comprehensible models. 

It is important that research in both interpretable and comprehensible systems continue forward. 
This is because,
depending on the user's background and her purpose of employing an AI model,
one type is preferable to another. As a real-life example of this,
most people think of a doctor as a kind of black box that transforms symptoms
and test results into a diagnosis. Without providing
information about the way medical tests and evaluations work, doctors  
deliver a diagnosis to a patient by explaining high-level indicators 
revealed in the tests (i.e. system symbols). 
Thus, when facing a patient, the physician should be like a comprehensible model.
When interacting with other doctors and medical staff, however, the 
doctor may be like an interpretable model: She can sketch a technical line of connecting patient symptoms and test
results to a particular diagnosis. Other doctors and staff can interpret a
diagnosis in the same way that an analyst can interpret an ML model, 
ensuring that the conclusions drawn are supported by reasonable
evaluation functions and weight values for the evidence presented.

\begin{figure}
\vspace{-15px}
\centering
\includegraphics[width=0.5\textwidth]{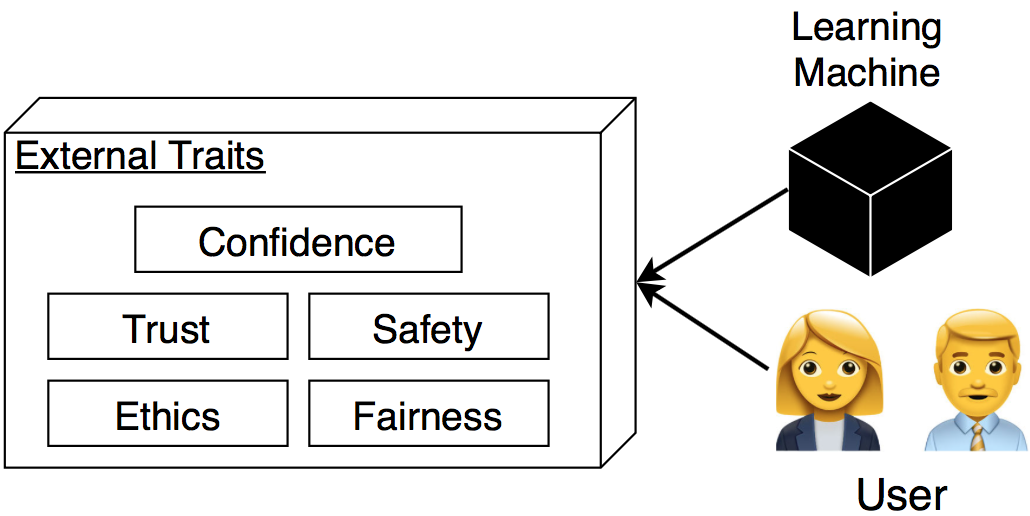}
\caption{External traits of a machine related to explainable AI. The traits
depend not only on properties of the learning machine, but also the user. For example,
confidence in an interpretable learning system is a function of the user's capability 
to understand the machine's input/output mapping behavior.} 
\label{fig:traits}
\vspace{-20px}
\end{figure}

{\bf Explainable system traits.}
Often discussed alongside explainable AI are the
external traits such systems should exhibit. These traits are 
seen as so important that some authors argue an AI system is not `explainable' if it does not support them~\cite{lipton16}.
Figure~\ref{fig:traits} presents such traits and conveys 
their dependence on not only the learning model but also the user. 
For example, explainable AI should
instill confidence and trust that the model operates accurately. 
Yet the perception of trust is moderated by a user's internal bias for or against
AI systems, and their past experiences with their use. Safety, ethicality, and fairness are traits that can only be evaluated by a
user's understanding of societal standards and by her 
ability to reason about emitted symbols or 
mathematical actions.
Present day systems fortunately leave this reasoning to the user, 
keeping a person as a stopgap preventing unethical or unfair 
recommendations from being acted upon.

We also note that ``completeness'' is not an explicit trait, and might not even be desirable as such. Continuing with the doctor example from above, it 
may be desirable for a system to present a simplified (in the sense of incomplete, as opposed to abstracted) `explanation' similar to a doctor using a patient's incomplete and possibly not entirely accurate preconceptions in explaining a complex diagnosis, or even sparing the patient especially worrisome details which might not be relevant for the subsequent treatment.
\vspace{-10px}

\section{Truly Explainable AI Should Integrate Reasoning}
\begin{figure}
\vspace{-15px}
\centering
\includegraphics[width=0.8\textwidth]{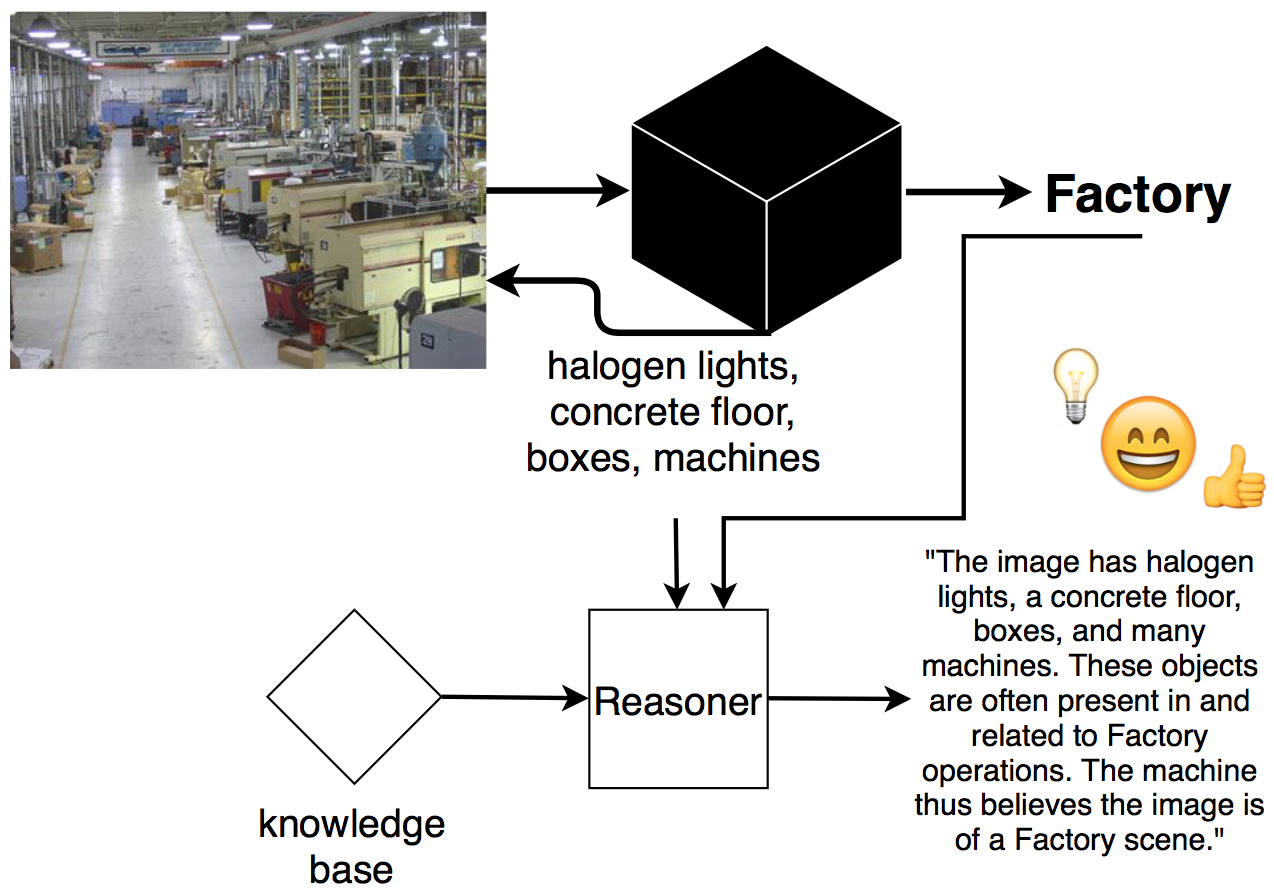}
\caption{Augmenting comprehensible models with a reasoning engine. 
This engine can combine symbols emitted by a comprehensible machine with a 
(domain specific) knowledge base encoding relationships between concepts 
represented by the symbols. The relationships between symbols in the knowledge 
based can yield a logical deduction about their relationship to the 
machine's decision. 
} 
\label{fig:neuro}
\vspace{-20px}
\end{figure}

\vspace{-10px}
Interpretable and comprehensible models encompass much of the present work
in explainable AI. Yet we argue that both approaches are lacking in their
ability to formulate, for the user, a line of {\em reasoning} that explains 
the decision making process of a model {\em using human-understandable
features of the input data}. Reasoning
is a critical step in formulating an explanation about why or how some 
event has occurred (see, e.g., Figure~\ref{fig:neuro}). Leaving explanation generation to 
human analysts can be dangerous since, depending on their background knowledge 
about the data and its domain, different explanations about why a model
makes a decision may be deduced. 
Interpretive and comprehensible models thus 
{\em enable} explanations of decisions, but do 
not yield explanations themselves. 

Efforts in neural-symbolic integration \cite{garcez2015} aim to develop methods which might enable explicit automated reasoning over model properties and decision factors by extracting symbolic rules from connectionist models. Combining their results with work investigating factors influencing the {\em human comprehensibility} of representation formats and reasoning approaches \cite{schmid2017} might pave the way towards systems effectively providing full explanations of their own to their users.

\noindent {\bf Acknowledgement.} The authors thank the Schloss Dagstuhl -- Leibniz Center for Informatics and organizers and participants of Dagstuhl Seminar 17192 
on Human-Like Neural-Symbolic Computing for providing the environment to develop 
the ideas in this paper. This work is 
partially supported by a Schloss Dagstuhl travel grant 
and by the Ohio Federal Research Network. Parts of the work have been carried out at the Digital Media Lab of the University of Bremen.

\bibliographystyle{splncs03}
\bibliography{cexai17}

\begin{thebibliography}{10}
\providecommand{\url}[1]{\texttt{#1}}
\providecommand{\urlprefix}{URL }

\bibitem{al06}
Al~Hasan, M., Chaoji, V., Salem, S., Zaki, M.: Link prediction using supervised
  learning. In: Wkshp. on link analysis, counter-terrorism and security (2006)

\bibitem{bau17}
Bau, D., Zhou, B., Khosla, A., Oliva, A., Torralba, A.: {Network Dissection:
  Quantifying Interpretability of Deep Visual Representations}. In: Proc. of
  Computer Vision and Pattern Recognition (2017)

\bibitem{garcez2015}
Garcez, A.d., Besold, T.R., De~Raedt, L., F{\"o}ldiak, P., Hitzler, P., Icard,
  T., K{\"u}hnberger, K.U., Lamb, L.C., Miikkulainen, R., Silver, D.L.:
  Neural-symbolic learning and reasoning: contributions and challenges. In:
  Proceedings of the AAAI Spring Symposium on Knowledge Representation and
  Reasoning: Integrating Symbolic and Neural Approaches, Stanford (2015)

\bibitem{gerber14}
Gerber, M.S.: {Predicting crime using Twitter and kernel density estimation}.
  Decision Support Systems  61,  115--125 (2014)

\bibitem{johnson16}
Johnson, J., Karpathy, A., Fei-Fei, L.: Densecap: Fully convolutional
  localization networks for dense captioning. In: Proc. of Computer Vision and
  Pattern Recognition. pp. 4565--4574 (2016)

\bibitem{karpathy15}
Karpathy, A., Fei-Fei, L.: Deep visual-semantic alignments for generating image
  descriptions. In: Proc. of Conference on Computer Vision and Pattern
  Recognition. pp. 3128--3137 (2015)

\bibitem{katti17}
Katti, H., Arun, S.: {Can you tell where in India I am from? Comparing humans
  and computers on fine-grained race face classification}. arXiv preprint
  arXiv:1703.07595  (2017)

\bibitem{kocabey17}
Kocabey, E., Camurcu, M., Ofli, F., Aytar, Y., Marin, J., Torralba, A., Weber,
  I.: {Face-to-bmi: Using computer vision to infer body mass index on social
  media}. arXiv preprint arXiv:1703.03156  (2017)

\bibitem{lipton16}
Lipton, Z.C.: The mythos of model interpretability. Workshop on Human
  Interpretability in Machine Learning  (2016)

\bibitem{maaten08}
Maaten, L.v.d., Hinton, G.: {Visualizing data using t-SNE}. Journal of Machine
  Learning Research  9,  2579--2605 (2008)

\bibitem{michie88}
Michie, D.: Machine learning in the next five years. In: Proc. of the Third
  European Working Session on Learning. pp. 107--122. Pitman (1988)

\bibitem{saradhi11}
Saradhi, V.V., Palshikar, G.K.: Employee churn prediction. Expert Systems with
  Applications  38(3),  1999--2006 (2011)

\bibitem{schmid2017}
Schmid, U., Zeller, C., Besold, T.R., Tamaddoni-Nezhad, A., Muggleton, S.: {How
  Does Predicate Invention Affect Human Comprehensibility?} Inductive Logic
  Programming: ILP 2016 Revised Selected Papers pp. 52--67 (2017)

\bibitem{shafiq14}
Shafiq, M.Z., Erman, J., Ji, L., Liu, A.X., Pang, J., Wang, J.: Understanding
  the impact of network dynamics on mobile video user engagement. In: ACM
  SIGMETRICS Performance Evaluation Review. pp. 367--379 (2014)

\bibitem{skirpan17}
Skirpan, M., Yeh, T.: {Designing a Moral Compass for the Future of Computer
  Vision using Speculative Analysis}. In: Proc. of Computer Vision and Pattern
  Recognition (2017)

\bibitem{you16}
You, Q., Jin, H., Wang, Z., Fang, C., Luo, J.: Image captioning with semantic
  attention. In: Proc. of Computer Vision and Pattern Recognition. pp.
  4651--4659 (2016)

\end{thebibliography}

\end{document}